\documentclass[letterpaper, 10 pt, conference]{ieeeconf} 

\IEEEoverridecommandlockouts                              

\usepackage{float}
\usepackage[dvipsnames]{xcolor}
\usepackage{graphicx}
\usepackage{multicol}
\usepackage{multirow}
\usepackage{tabularx}
\usepackage{subfig}
\usepackage{svg}
\usepackage{amsmath}
\usepackage{amsfonts}
\graphicspath{./images/}
\usepackage{cite}
\usepackage{array}
\usepackage{algorithm}
\usepackage{algorithmic}

\title{\LARGE \bf
A Twin Delayed Deep Deterministic Policy Gradient Algorithm for Autonomous Ground Vehicle Navigation via Digital Twin Perception Awareness*
}

\author{Kabirat Olayemi$^{1}$, Mien Van$^{1}$, Sean McLoone$^{1}$, Yuzhu Sun$^{1}$, Jack Close$^{1}$, Nguyen Minh Nhat$^{1}$, and Stephen McIlvanna$^{1}$ 
\thanks{*This work was supported by The Department for Education, Northern Ireland.}
\thanks{$^{1}$The authors are with The School of Electronics, Electrical Engineering and Computer Science,
        Queens University, Belfast, UK
        {\tt\small \{kolayemi01, m.van, s.mcloone, ysun32, jclose06, nnhat01, smcilvanna01\}@qub.ac.uk.}}%
}

\begin{document}

\maketitle
\thispagestyle{empty}
\pagestyle{empty}

\begin{abstract}

Autonomous ground vehicle (UGV) navigation has the potential to revolutionize the transportation system by increasing accessibility to disabled people, ensure safety and convenience of use. However, UGV requires extensive and efficient testing and evaluation to ensure its acceptance for public use. This testing are mostly done in a simulator which result to sim2real transfer gap. In this paper, we propose a digital twin perception awareness approach for the control of robot navigation without prior creation of the virtual environment (VT) environment state. To achieve this, we develop a twin delayed deep deterministic policy gradient (TD3) algorithm that ensures collision avoidance and goal-based path planning. We demonstrate the performance of our approach on different environment dynamics. We show that our approach is capable of efficiently avoiding collision with obstacles and navigating to its desired destination, while at the same time safely avoids obstacles using the information received from the LIDAR sensor mounted on the robot. Our approach bridges the gap between sim-to-real transfer and contributes to the adoption of UGVs in real world. We validate our approach in simulation and a real-world application in an office space. 

\end{abstract}

\section{INTRODUCTION}

Autonomous driving has the potential to revolutionise the transportation system in the near future. One of the primary promises of autonomous driving is improved accessibility to people who cannot drive due to age or disability, among other future prospects such as improved safety, convenience of use, and optimised traffic flow. 

Public acceptance of autonomous driving technology requires rigorous testing and validation of the technology, especially path planning and odometry control tasks \cite{hu2023simulation}. Recently, the application of reinforcement learning (RL) and deep reinforcement learning (DRL) in autonomous driving to generate a control strategy has shown better performance control \cite{voogd2023reinforcement}. Developing a DRL control strategy for autonomous driving requires exploring its environment several times for optimal performance and ensuring safety and reliability, which is impossible without the help of simulators \cite{9499331}. Real-life training is unsafe, time-consuming, and expensive to use. 

Simulators are used for the development, testing, and validation of autonomous driving systems, providing a safe, cost-effective, and scalable environment. Once the model is developed and validated, it is deployed in the real environment \cite{8450505}. However, simulated environments may not fully replicate the complexity and unpredictability of the real world. Factors such as human behavior, infrastructure variability, and dynamic road conditions can be challenging to accurately simulate, leading to potential discrepancies between simulated and real-world performance. Also, simulators rely on models to simulate sensor inputs such as LiDAR, radar, and cameras. These models may not capture the full range of sensor behaviours and limitations encountered in real-world conditions, potentially leading to inaccuracies in perception and decision-making by autonomous systems.

To close the gap between the simulator and the real world environment (SRE), different methods have been proposed \cite{9308468}. The simulation-to-reality transfer technique fine-tunes learnt policies or models using real-world data, using techniques such as domain adaptation, transfer learning, or imitation learning to improve performance on real-world tasks\cite{kang2019generalization, liu2020real, ju2022transferring}
. However, this method presents several challenges, such as: (i) Uncertainties in simulation models can lead to discrepancies in behaviour between simulated and real-world scenarios, challenging the generalisation of learned policies to real-world environments. (ii) Achieving sim-to-real transfer with high-fidelity simulations and complex models can be computationally intensive and resource-demanding.

Another method adopted to close SRE is the digital twin (DT). DT is a virtual representation of a physical object, system, or process that provides a real-time, data-driven simulation of its behavior and performance. In the context of autonomous driving, a digital twin can represent various components of the vehicle\cite{dygalo2020principles}, the surrounding environment\cite{wu2021digital}, and the interactions between them\cite{9869610}.

Start-of-the-art DT autonomous driving focus is on the driver assistance system, the traffic avoidance system, and the control system. However, these approaches have two shortcomings: (i) the representation of the virtual twin (VT) is pre-created and training is based on the created gazebo environment. (ii) The mapping of the environment state between the digital world and the virtual world updates the experience buffer but does not allow retraining of the model in real time.

\begin{figure*}
    \centering
    \includegraphics[width=1\linewidth]{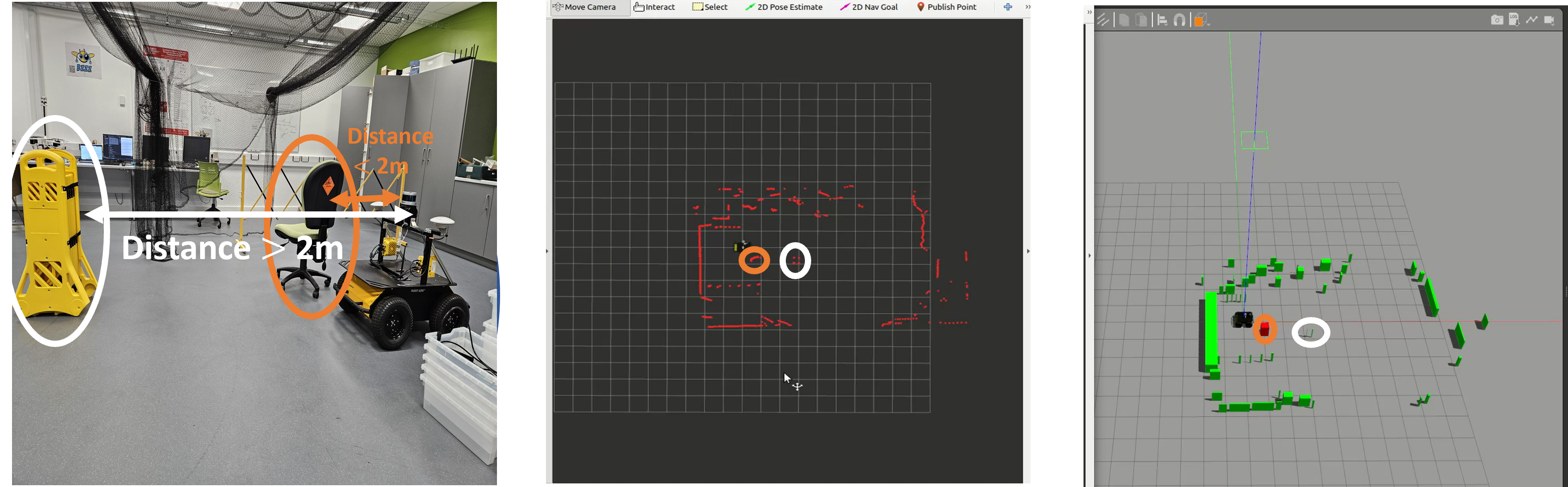}
    \caption{Illustration of the VT gazebo environment creation based of the information received from the  LIDAR sensor pointcloud data. The leftmost figure shows the environment and it's corresponding poincloud representation in RVIS is shown in the middle figure. The leftmost figure VT of the created from the processed LIDAR data. The robot shown is an unmanned ground vehicle (HUSKY A200) developed by ClearPath. In the gazebo, all obstacles at a distance greater than 1m from the robot are represented in green, while obstacles less than $2m$ from the robot are in red.}
    \label{fig_gazebo}
\end{figure*}

In this study, we propose a LIDAR-based perception of the environment state for the control of robot navigation without prior creation of the VT environment state. Our approach plans a path to a desired goal, while at the same time safely avoids obstacles using the information received from the LIDAR sensor mounted on the robot. We achieve this by developing a twin delayed deep deterministic policy gradient (TD3) algorithm that ensures collision avoidance and goal-based path planning. Unlike existing methods that pre-create the gazebo environment for training the TD3 model, our method creates a VT representation of the environment based on the information received from the physical LIDAR sensor pointcloud as shown in Figure \ref{fig_gazebo}.

To create VT, we propose a novel perception method that preprocesses the pointcloud data into clusters of close neighbours to extract properties such as centroid (x,y position) and size of obstacles in the world. The received obstacle properties are then spawned in the gazebo environment for training. This way, we overcome the problem of pre-creating of the VT gazebo and the VT can change whenever the physical twin (PT) changes. 

To control the navigation of the physical husky, we developed connection between twin. The VT controls the movement of the PT. The VT pauses the movement of the PT robot once the VT robot is within a threshold set, as the danger zone then retrains until a path is found. Once a path is found, the model is updated and the VT checks if the state of the robot has changed and safe to unpause the PT robot to continue navigating. In this way, we can enable retraining of the TD3 network when the environment dynamic becomes more complex and the existing model could not navigate safely. 

We demonstrate the performance of our approach on different environment dynamics. We show that our approach is capable of efficiently avoiding collision with obstacles and navigating to its desired destination. We intend that our approach bridges the gap between sim-to-real transfer and contribute to the adoption of UGVs in real world. In conclusion, our contributions are summarised as follows:
\begin{enumerate}
    \item A digital twin that controls and plans UGV navigation system.
    \item A novel perception method that creates the gazebo environment of the VT based on the state of the PT environment.
    \item A model that enables online retraining of the TD3 network model.
\end{enumerate}
The paper is organised as follows. Section \ref{sec_literature} reviews previous related work on reinforcement learning for collision avoidance and path planning. Section \ref{sec_Background} reviews the basic theory of simulators and digital twin. The experimental setup adopted in this study is discussed in \ref{sec:methodology}. Section \ref{sec:results} discusses the experimental results and the performance obtained in this study. Finally, Section \ref{sec:conclusion} concludes the article and states our future direction.

\section{Related Work}
\label{sec_literature}
Existing path-planning and collision avoidance methods is discussed in \cite{mohammed2020perception,guastella2020learning}. Model-based predictive control (MPC) is a control method that uses a dynamic model of the system being controlled to predict its future behaviour and optimise control actions over a specified time horizon \cite{schwenzer2021review}. MPC has been used for collision avoidance in a number of works \cite{hang2021path, xue2020active, hu2020steering, kouvaritakis2021stochastic} because it provides a flexible and robust framework for collision avoidance, allowing vehicles or robots to navigate safely in dynamic environment. However, with MPC, its computational complexity is relatively high and additional effort is required to handle modelling errors and disturbances. 

Control barrier function (CBF) is another main method for collision avoidance. It provides a framework for designing control policies that guarantee the satisfaction of safety constraints over a finite time horizon. A well-designed CBF can ensure safety in a static environment or in a known dynamic environment \cite{yu2023sequential}. However, while CBFs offer a principled approach to collision avoidance and safety-critical control, it's application in dynamic environment is challenging. CBFs tend to be conservative in nature, which means that they may prioritise safety at the expense of performance or efficiency. This conservatism can lead to overly cautious behaviour, such as excessive braking or maneuvering. Also, designing and implementing CBF-based control policies can be computationally intensive, particularly in dynamic or uncertain environments. Hence, CBFs are applied with other methods \cite{singletary2021comparative, zhang2023neural, thontepu2022control} to mitigate these limitations. 

DRL is a subset of machine learning and artificial intelligence that combines deep learning techniques with reinforcement learning principles to enable agents to learn optimal action by interacting with its environment through trial and error. Many current research on autonomous driving attempts to adopt various deep reinforcement learning algorithms for path planning and collision avoidance, such as Deep Q-learning (DQN) \cite{mohanty2020context,zhai2022intelligent}, Deep Deterministic Policy Gradient (DDPG) \cite{chen2022target}, Soft-Actor Critic (SAC) network \cite{haarnoja2018soft}, and Twin-delay Deep Deterministic Policy Gradient (TD3)\cite{yang2022automatic}. These DRL algorithms are a powerful paradigm for learning decision-making policies in simple, complex, static, or dynamic environments. However, DRL algorithms often require large amount of interaction with the environment to learn effective policies. In addition,ensuring that DRL models generalise well to an unknown environment is an ongoing research challenge.    

\section{Background}
\label{sec_Background}
\textit{Simulators:} Due to recent advances in DRL algorithms for autonomous vehicle control, simulators have been widely used in academia and industry for research, development and testing of autonomous driving algorithms and systems \cite{dosovitskiy2017carla}. In simulators, researchers and engineers can implement the dynamics required for real-world driving, such as vehicle configuration, environment complexity, traffic laws, and sensor configuration and positioning. Several different simulators have been used, each having its strengths and may be preferred based on specific project requirements and constraints, see Table \ref{tab:simulators}. Open source simulators are freely available for researchers to use and customise to their needs, unlike commercial simulators, which come with a price and cannot be easily adapted to users' requirements. 
\begin{table*}
    \centering
    \begin{tabular}{|p{2cm}|p{2cm}|p{3.5cm}|p{3.5cm}| p{3.5cm}|}
\hline
\textbf{Simulators} & \textbf{Developed} & \textbf{Features} & \textbf{Performance}& \textbf{Limitation} \\
\hline
\textbf{CarSim} \cite{jiang2022modelling} & Mechanical Simulation Corporation   & Support for co-simulation with other software tools, such as Simulink, to incorporate control algorithms. Realistic simulation of vehicle manoeuvres, such as brakes, cornering, and lane changes. & High accuracy in simulating vehicle dynamics, making it suitable for detailed analysis and optimisation of vehicle performance.&It is not open source and hence costly and lacks some features such as high-fidelity sensor simulation and scenario generation. \\
\hline
\textbf{CARLA(Car Learning to Act)} \cite{deschaud2021kitti}  &  Intel and Toyota Research Institute & Open-source, Python and C++ API, traffic simulation, urban environment simulation   & Versatile and good for research and industry & limited support for high-fidelity sensor simulation. The fidelity of sensor simulation may not match real-world conditions. \\ 
\hline
\textbf{AirSim} \cite{shah2018airsim}  & Microsoft   & Open-source, supports various environments urban or rural, provides APIs for C++ and Python  & Flexible and easy integration with ML frameworks like Tensorflow and PyTorch. & Complexity of setup and configuration. Also, it has limited support for real-time simulation. \\ 
\hline
\textbf{LGSVL} \cite{rong2020lgsvl}   & LG Electronics   & Unity-based, supports urban environment, supports various sensors, provides Python and C\# APIs. & Offers a user-friendly interface and facilitates testing of autonomous driving algorithms. & Integrating custom sensors, control algorithms, or scenario generation techniques may require additional effort and expertise.  \\
\hline
\textbf{WODS (Waymo Open Dataset Simulation)} \cite{madrigal2017inside} & Waymo  & Part of the Waymo Open Dataset initiative, designed for large-scale simulation of autonomous driving scenarios.   & Allows training and evaluating models in simulation before deploying them in real-world settings. & Dependency on the Waymo Open Dataset which may limit its flexibility and generalizability to other datasets and environments. \\ 
\hline
\textbf{Gazebo} \cite{7521461}  & Open Source Robotics Foundation.   & Open-source, widely used in robotics and autonomous systems, supports sensor simulation, can be extended using plugins, often used with the Robot Operating System (ROS). & Recognized for its versatility in simulating robotic systems, including autonomous vehicles. & Limited support for high-fidelity sensor simulation and difficult to customise Gazebo for specific autonomous driving scenarios or integrating external components. \\
\hline
    \end{tabular}
    \caption{Simulators for Autonomous Driving}
    \label{tab:simulators}
\end{table*}

One of the most widely used simulators is the Gazebo, developed by Open Source Robotic Foundation \cite{mingo2014yarp,furrer2016rotors}. Gazebo is free, supports Robotic Operating System (ROS), and offers the ability to simulate an indoor, outdoor, static, or dynamic environment. The robot model and environment dynamic are defined in a .URDF (Universal Robot Description Format and .world file respectively. The URDF and world files support XML (Extensible Markup Language) and xacro (XML macro) languages which are easy to read, implement and update \cite{9311330}. Another main feature of Gazebo is that it supports wired and wireless network communication \cite{9213892}, allowing the sharing of information between multiple robots or platforms. Though, Gazebo has limited support for simulating high-fidelity sensors, in this paper, we will introduce a DT training where the real-world sensor is passed over a network to the Gazebo as input for training a TD3 model.

\textit{Husky A200:} Husky A200 is a robust mechanical four-wheel drive vehicle capable of traversing rough terrain, uneven surfaces, and outdoor environments with obstacles. It is compatible with ROS (Robot Operating System), providing a flexible and modular software framework for development, control, and integration of robotic applications. It supports Ethernet, Wi-Fi, and other communication interfaces for remote operation and data exchange.

\textit{Digital Twin:} A digital twin is a virtual representation of a real-world object, system, or process. It is a digital counterpart that mirrors the physical entity in terms of its characteristics, behaviour, and state \cite{liu2023digital, wu2021digital}. DT was introduced by the National Aeronautics and Space Administration (NASA) Apollo programme in the 1970s. They built a replica of space vehicles on Earth to mirror the condition of the equipment during the mission \cite{hu2021digital}. Since then, it has been widely used in various industries for simulation, analysis, monitoring, and control purposes. 

\begin{figure}
    \centering
    \includegraphics[width=1\linewidth]{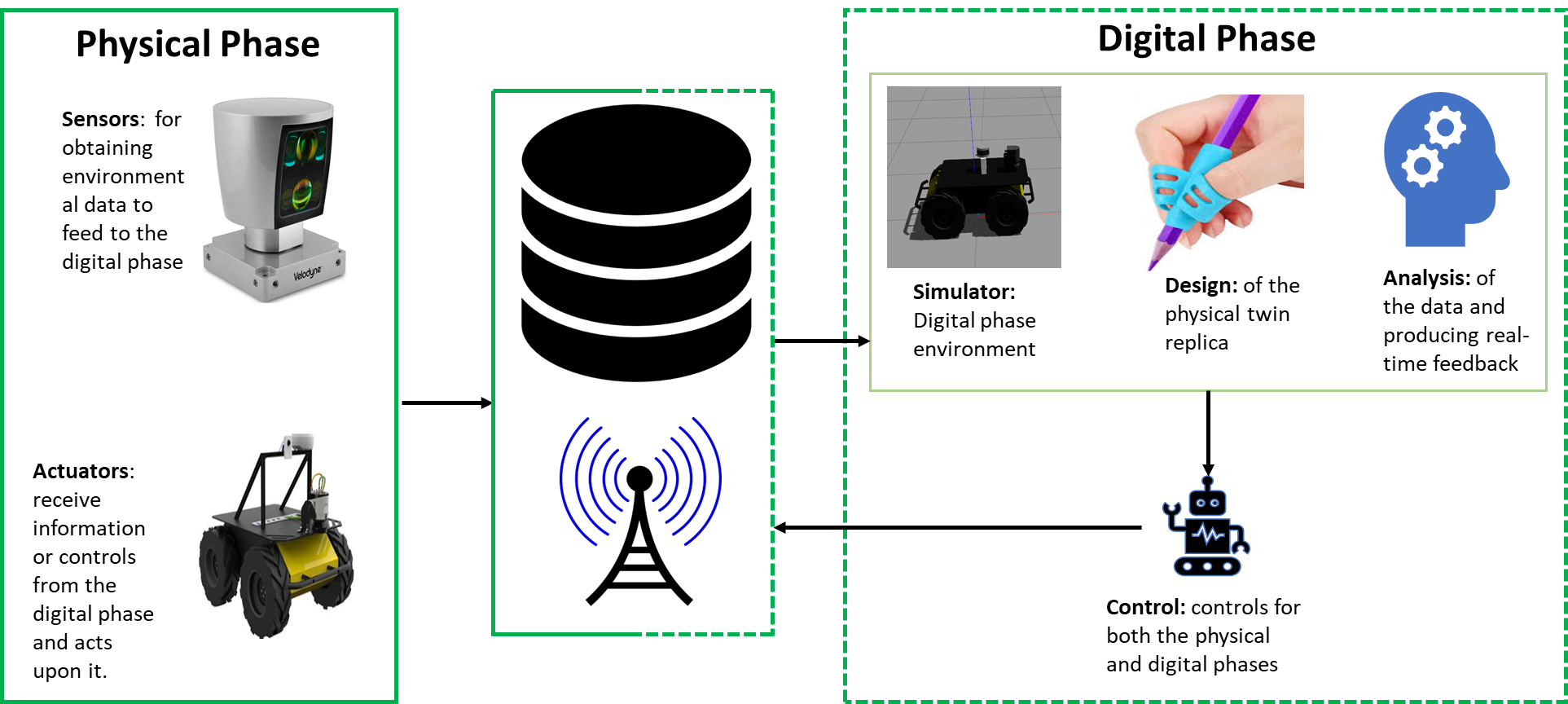}
    \caption{Digital twin key components for real-time data integration.}
    \label{fig_DT_framework}
\end{figure}

DT key components are physical space, virtual space, connection or network, and data storage as shown in Figure \ref{fig_DT_framework}. The physical space basically consists of sensors and actuators. The sensor collects data from various sources, such as Lidar, IoT devices, historical databases, and external systems for feeding to the digital space while the actuators receives process feedback from the digital space. The digital space models the representation of the physical space, receives data or changes from the physical space for analysis and feedback controls to the actuators. In between the physical space and the digital space, there exists a database for data fusion and storage, and a network for connection between the phases. 

\section{Methodology}
\label{sec:methodology}
\begin{figure}
    \centering
    \includegraphics[width=1\linewidth]{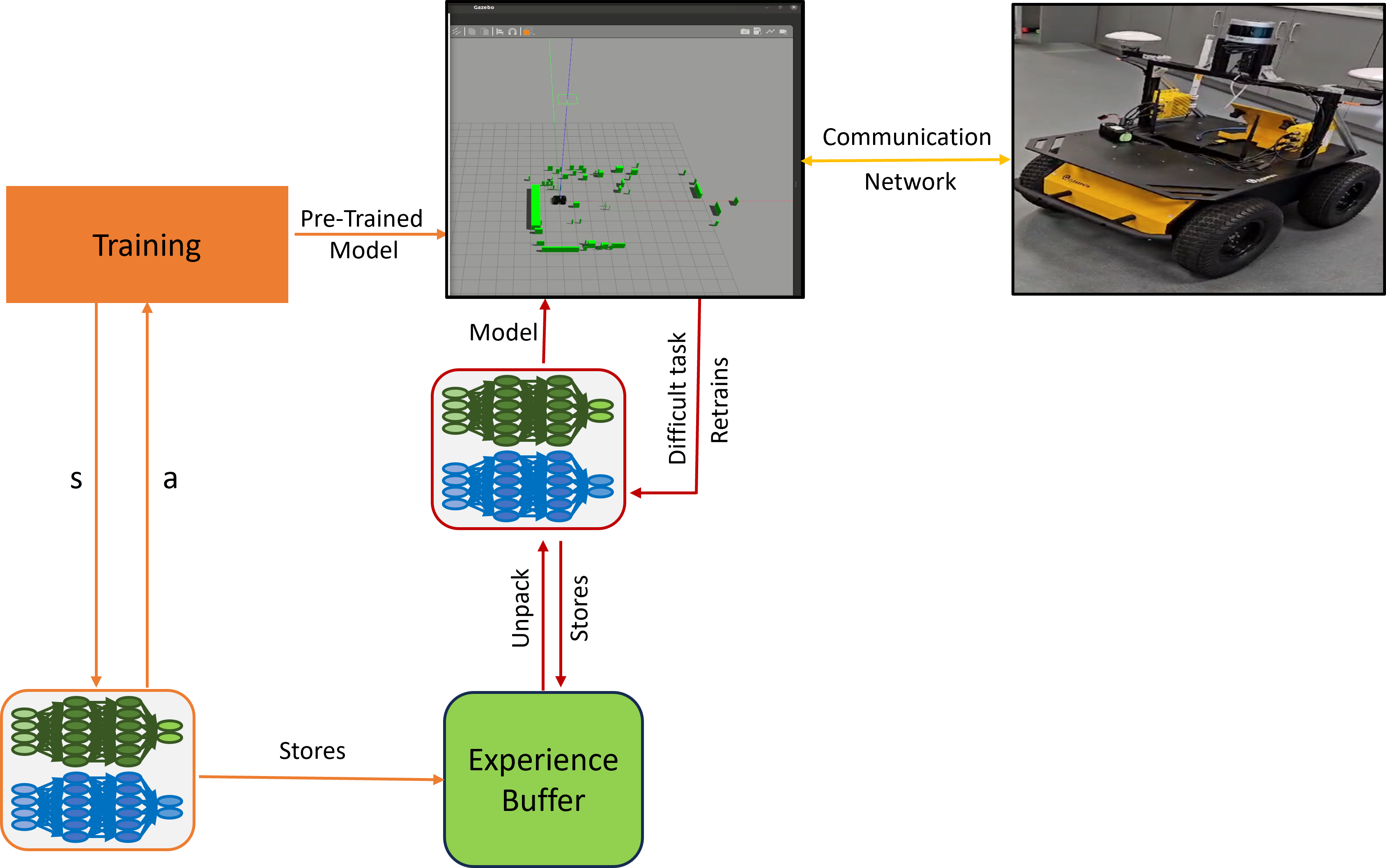}
    \caption{Digital twin key components for real-time data integration.}
    \label{fig_method_overview}
\end{figure}

This section discusses the proposed algorithm in designing the digital twin autonomous navigation policy. In this paper, we consider developing an initial pre-trained TD3 model for self-navigation using gazebo simulator then a digital twin is established to test the performance of the model. We re-initiate a retraining if there exist a complex environment the robot could not navigate based on the existing model. The overview of our digital twin is shown in Figure \ref{fig_method_overview}. 

\subsection{\textit{ Reinforcement Learning Pre-trained Algorithm}}
Generally, a learning-decision making policy can be formulated as a Markov Decision Process (MDP), which is defined by the state space $(S)$, the action space $(A)$, the state transition probability $(P)$, the reward function $(R)$, and the discount factor $(\gamma)$. 
\subsubsection{State Space}
The state space is a set of spaces representing all possible situations or configuration. We define our state as $S = [l_n, \theta, g, r, v, w]$, where $l = {1,2,...,n}$ represent the LIDAR sensor data received from the robot, $\theta$ is the orientation of the robot, $g$ is the goal position, $r$ is the robot position, $v$ is the linear velocity of the robot, and $w$ is the angular velocity of the robot. 

\subsubsection{Action Space} 
At each state, the agent chooses from a possible set of actions it can take. The available actions are dependent on the model of the agent, which in turn influences the outcome of the decision-making process. For husky being a differential drive robot, it's possible action $A = [v,w]$ are the linear speed $v$ and angular speed $w$. 

\subsubsection{Transition Probability}
The transition probability is the probability of transitioning from one state to another after taking a specific action $P = (s_{t+1}|s_t,a_t)$. 

\subsubsection{Reward Function}
Rewards represent an immediate feedback received by an agent after taking an action in a particular state. The reward can be positive or negative. A positive reward is given to the agent for reaching its target (\ref{eq_goal}), negative reward is given if the robot collides with an obstacle (\ref{eq_col}) and if the robot neither collides with an obstacle nor reaches its target, it receives an intermediate reward \ref{eq_intermediate} based on the it's distance to the target (\ref{eq_dist}), the action it took (\ref{eq_action}), and it's orientation towards the goal (\ref{eq_orient}).

\begin{equation}\label{eq_goal}
R= 100
\end{equation}

\begin{equation} \label{eq_col}
R= -100
\end{equation}
\begin{equation}\label{eq_intermediate}
R= d + a + o
\end{equation}
\begin{equation}\label{eq_dist}
    d= \left\{
    \begin{array}{ll}
         \frac{1 - \text{distance}}{2}, & \mbox{if distance $< 1$}  \\
         0, & \mbox{otherwise} 
    \end{array}
    \right.
\end{equation}
\begin{equation}\label{eq_action}
a= \frac{a_l}{2} - \frac{|a_w|}{2}
\end{equation}

\begin{equation}\label{eq_orient}
o = \Vec{o} \cdot \frac{\Vec{t} - \Vec{r}}{\lVert \Vec{t} - \Vec{r} \rVert} \times 50
\end{equation}

where $\Vec{t}$ is the target position, $\Vec{r}$ is the robot's position and $\Vec{o}$ is the orientation vector of the robot.
\subsubsection{Discount Factor}
The discount factor ($\gamma = [0,1]$) is used to control the agent to value the future reward rather than the immediate rewards. 

\begin{figure}
    \centering
    \includegraphics[width=1\linewidth]{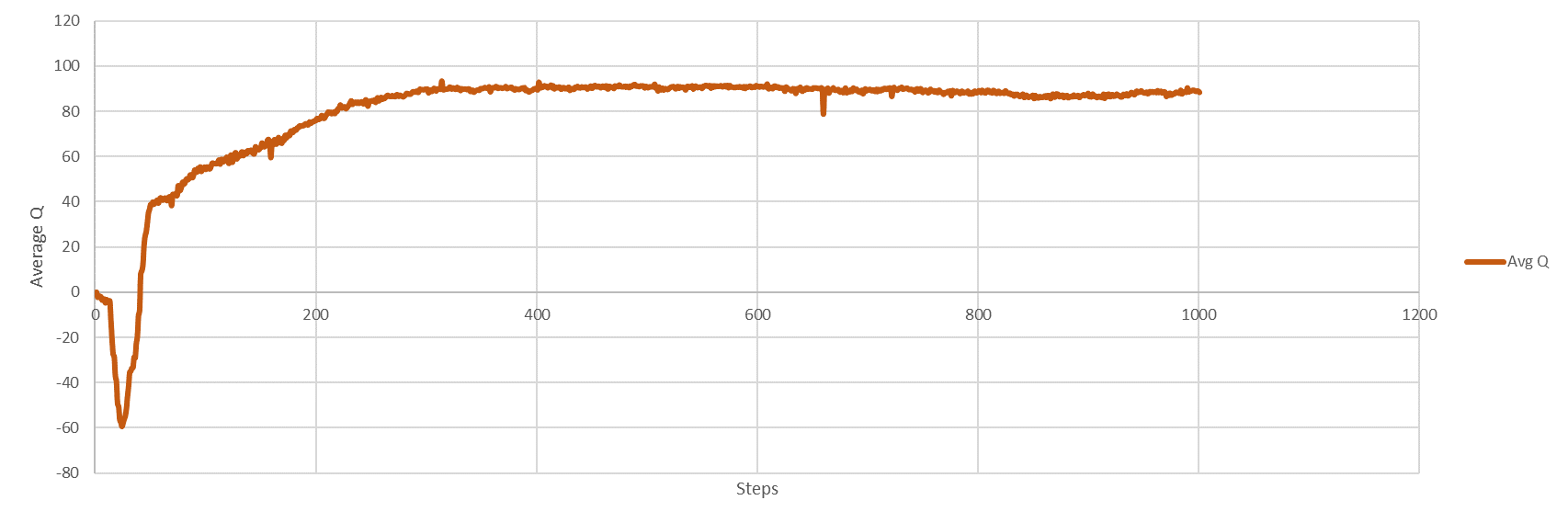}
    \caption{Digital twin key components for real-time data integration.}
    \label{fig_avgQ}
\end{figure}

The result of the pre-trained model shown in Figure \ref{fig_avgQ} shows that the average action value function of the TD3 network is gradually increased. This increase indicates that the network exhibits a better approximation of the true Q-value function.

\subsection{\textit{Digital Twin}}
We build the virtual environment to predict the state transition of the robot in the physical environment. As shown in Figure \ref{fig_method_overview}, the virtual environment and the physical environment share information over the network. The robot in the physical environment (PR) acting as the server receives linear and angular velocity values from the robot in the virtual space (VR) while VR acting as the client sends an initial $0.0$ velocities to the PR in other to receive LIDAR sensor data from PR to create a twin of the environment. 

The twin is created by processing the received LIDAR sensor data. Initially, we extract range data from the LaserScan message and convert it to a numpy array and replace infinite values with 0. Using the range data and the corresponding angles, we calculate the points in a 2D space based on polar coordinates. Then we applied the DBSCAN clustering algorithm to cluster the points as described in algorithms \ref{alg:dbscan},\ref{alg:regionquery}, and \ref{alg:expandcluster}. For each of the extracted clusters, we calculate its bounding box to determine its size (width and height) and position (centre point). Also, we calculate the distance from the robot's position to the cluster's centre using the Euclidean distance. The obtained size, position, and distance of the objects is the physical environment and is then spawned into the virtual space to create the representation of the physical space. Once the environment is created, the VR use the pre-trained model to control both itself and the PR. 
\begin{algorithm}
\caption{DBSCAN Algorithm}
\label{alg:dbscan}
\begin{algorithmic}[1]
\STATE \textbf{Input:} Set of points $P$, radius of neighborhood $\varepsilon$, minimum number of points $minPts$
\STATE \textbf{Output:} Clusters $C$
\FORALL{$p$ in $P$}
    \IF{$p$ is unvisited}
        \STATE Mark $p$ as visited
        \STATE $N \leftarrow$ \textsc{RegionQuery}($p$, $P$, $\varepsilon$)
        \IF{$|N| \geq minPts$}
            \STATE $C' \leftarrow$ \textsc{ExpandCluster}($p$, $N$, $P$, $\varepsilon$, $minPts$)
            \STATE Add $C'$ to $C$
        \ELSE
            \STATE Mark $p$ as noise
        \ENDIF
    \ENDIF
\ENDFOR
\end{algorithmic}
\end{algorithm}

\begin{algorithm}
\caption{RegionQuery Function}
\label{alg:regionquery}
\begin{algorithmic}[1]
\STATE \textbf{Input:} Point $p$, set of points $P$, radius $\varepsilon$
\STATE \textbf{Output:} Set of neighboring points $N$
\STATE $N \leftarrow \emptyset$
\FORALL{$q$ in $P$}
    \IF{\textsc{Distance}($p$, $q$) $\leq \varepsilon$}
        \STATE Add $q$ to $N$
    \ENDIF
\ENDFOR
\RETURN $N$
\end{algorithmic}
\end{algorithm}

\begin{algorithm}
\caption{ExpandCluster Function}
\label{alg:expandcluster}
\begin{algorithmic}[1]
\STATE \textbf{Input:} Core point $p$, neighboring points $N$, set of points $P$, radius $\varepsilon$, minimum number of points $minPts$
\STATE \textbf{Output:} Cluster $C$
\STATE $C \leftarrow \{p\}$
\FORALL{$q$ in $N$}
    \IF{$q$ is unvisited}
        \STATE Mark $q$ as visited
        \STATE $N' \leftarrow$ \textsc{RegionQuery}($q$, $P$, $\varepsilon$)
        \IF{$|N'| \geq minPts$}
            \STATE $N \leftarrow N \cup N'$
        \ENDIF
    \ENDIF
    \IF{$q$ is not yet a member of any cluster}
        \STATE Add $q$ to $C$
    \ENDIF
\ENDFOR
\RETURN $C$
\end{algorithmic}
\end{algorithm}
  
\subsection{\textit{Reinforcement Learning Re-training Algorithm}}
To bridge the gap between sim-to-real transfer, we implement a re-training mechanism to our approach. The pre-trained model controls the movement of both the VR and PR. If there exists any difficult task that the existing model could not solve, the movement of the physical robot is halted while the VR starts retraining and exploring it's environment to obtain a path and avoid collision. To save the exploration time, we load the pre-trained model, unpack experiences, and then continue training. Once a path is found in the virtual twin, the VR returns to the same point the retraining was initiated before sending controls to the PR.  

\section{Results}
\label{sec:results}
Our approach was implemented on an Intel Core i7-6800 CPU desktop computer with 32 GB RAM and NVIDIA GTX 1050 graphics card. We use Ubuntu 20.04 operating system and ROS Noetic 
 as both support Gazebo simulator for the experiment. We used Husky A200 in both simulation and real-life experiment. 
 
 We evaluated the performance of our approach using the metrics adopted in \cite{8733621}, which are the success rate, the collision rate, and the timeout rate.  The results of the test scenarios are listed in Table \ref{tab:result}. Our digital twin retraining approach shows the potential of bridging the gap between simulator to real transfer as it outperforms the traditional approach. 
\begin{table}
    \centering
    \begin{tabular}{cccc}
        \textbf{Method} & \textbf{Success} & \textbf{Collision}  & \textbf{Timeout} \\
    \hline
        Simulator test & 0.86 & 0.1 & 0.04 \\
        \hline
        Sim2real& 0.7 & 0.2 & 0.1\\
        \hline
        \textbf{Digital twin(ours)} & \textbf{0.98} & \textbf{0.02} & \textbf{0} \\
        \hline
    \end{tabular}
    \caption{Quantitative evaluation of the performance of the proposed digital twin collision avoidance and basic approaches}
    \label{tab:result}
\end{table} 
\subsection{Simulation Performance} 
\begin{figure}
    \centering
    \includegraphics[width=1\linewidth]{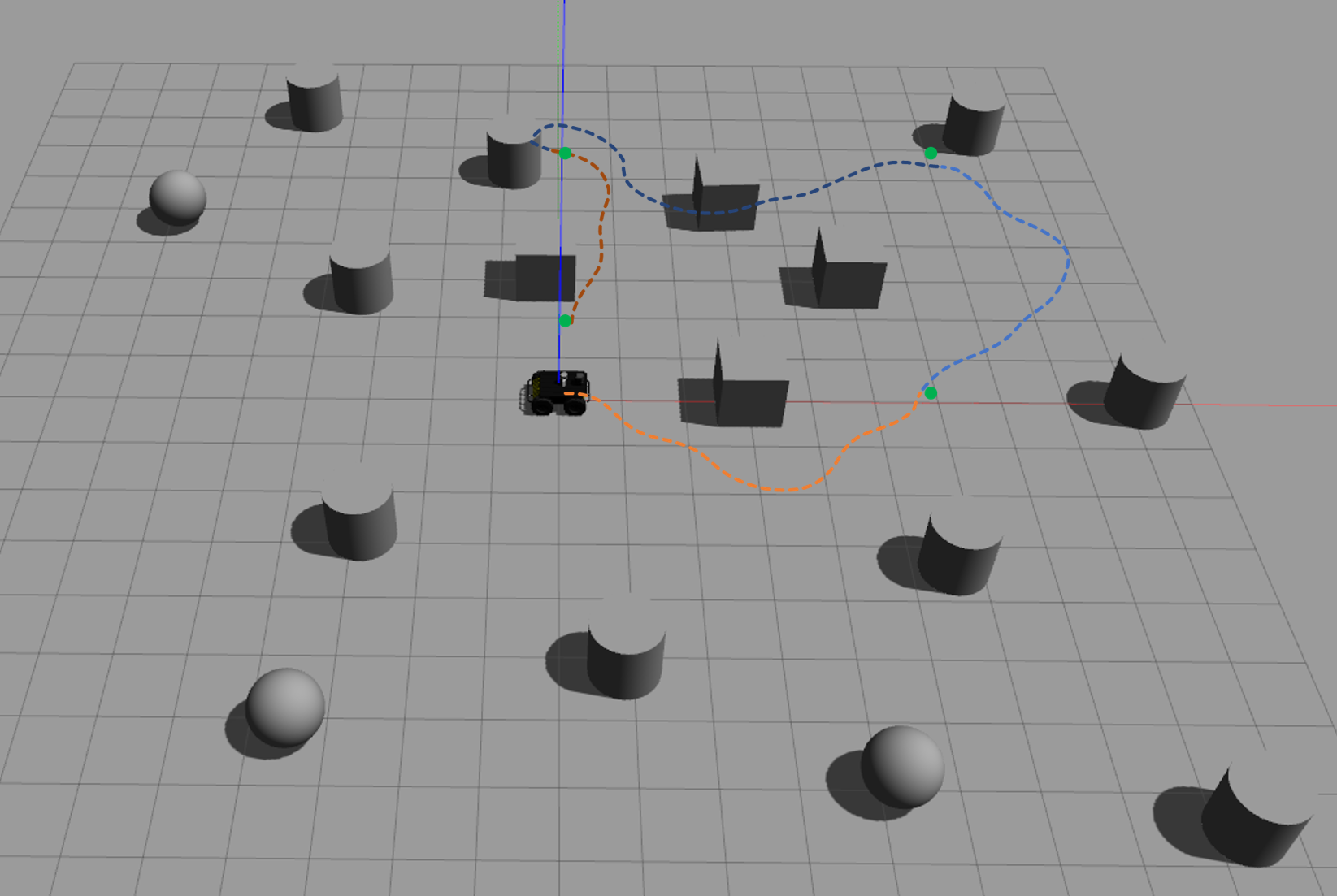}
    \caption{Sample of trajectories generated in the simulation environment. The initial position of the robot is 0.0 on both x and y axis for the initial test scenarios  while for other test scenarios it is the goal of the previous test scenarios. Each green point represents the target the robot navigated to, while the broken lines represent the path to each goal.}
    \label{fig:simulation_performance}
\end{figure}

Figure \ref{fig:simulation_performance} shows some of the trajectories in the gazebo simulator. Due to the width of husky, we allow a safe distance of $0.5m$ between the robot and the obstacles. Once the distance is below the defined threshold, it assumes that a collision has occurred. Hence, in the simulator, we sparsely distribute the obstacles ensuring that there is at least $1.2m$ between two obstacles. The result shows that the model can effectively navigate an environment with sparely distributed obstacles. This is so because we only trained the DRL model to learn how to navigate to it's intended target and simply avoid sparsely distributed static obstacles. We intend to establish a future retraining mechanism. 

\subsection{Performance without Digital Twin Retraining}\label{subsec: withou_retraining}
\begin{figure}
    \centering
        \begin{tabular}{cc}
        \hspace{-4pt}
           \includegraphics[width=4cm, height=6cm]{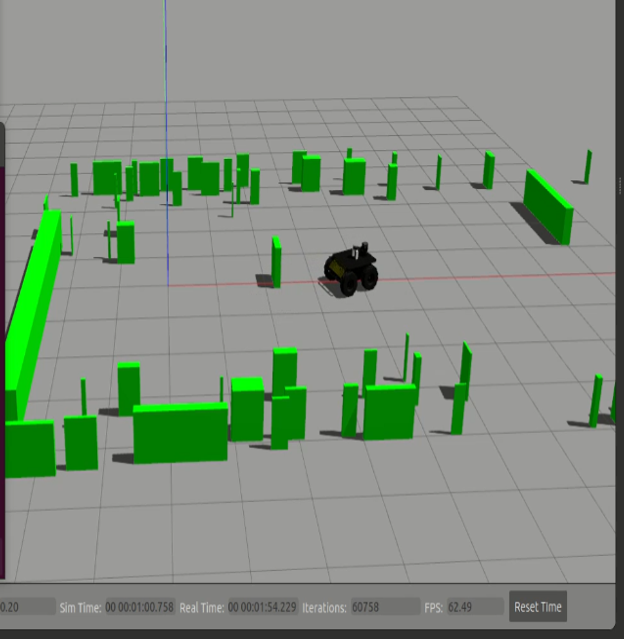}  & \includegraphics[width=4cm, height=6cm]{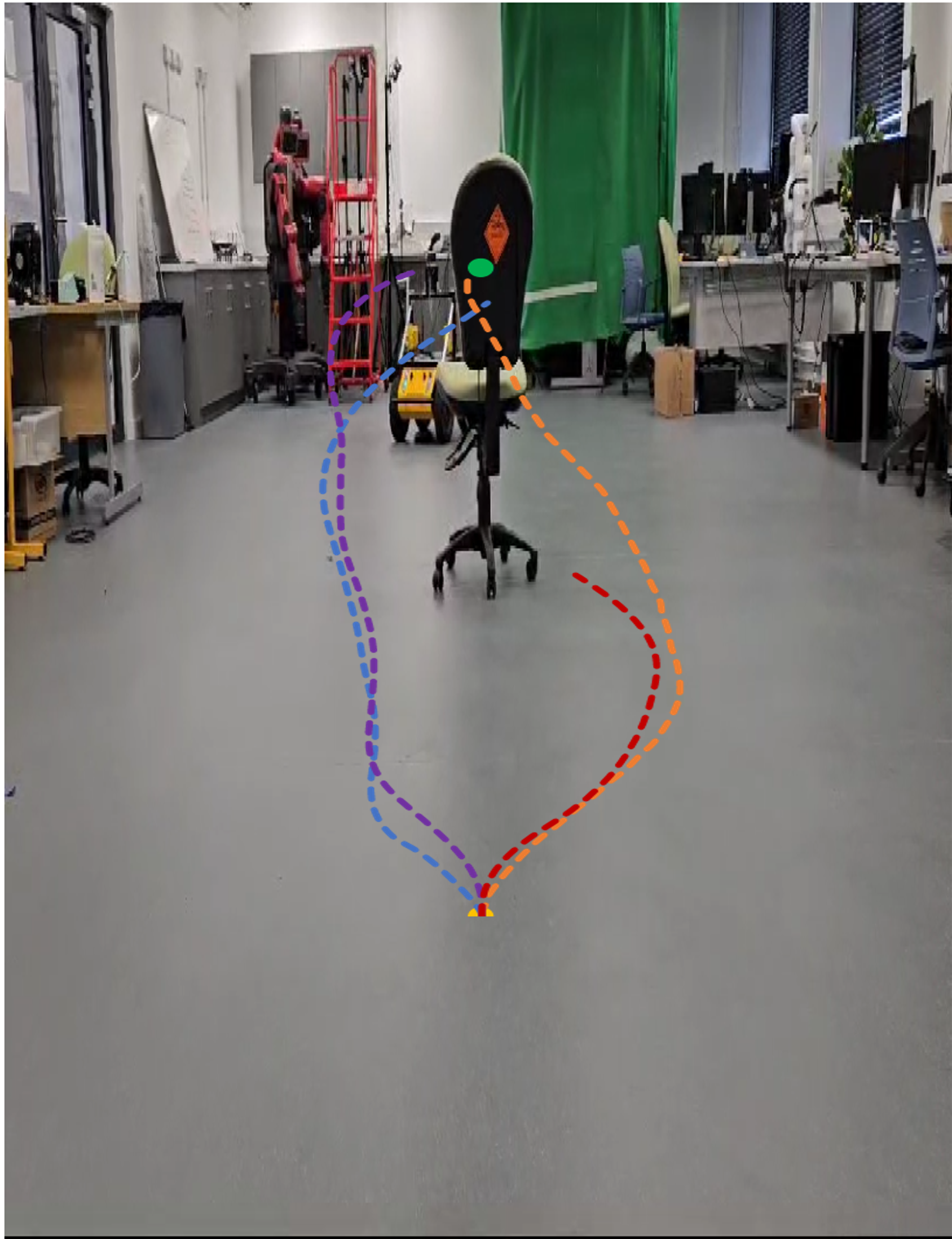}\\
           ({\bf a}) &({\bf b}) \\
        \end{tabular}
        \caption{(a) shows the virtual twin of the real robot state in the simulated environment. For each test case, the robot navigates from an initial position of 0,0 on both the x and y axes to a goal point of 5,0 on the x and y axes. Fig. (b) shows the path from the initial point to the goal point in the real world. The red coloured path indicates a collision while other paths indicate that the robot was able to drive to its goal point.}
        \label{fig:dt_performance}
\end{figure}

To establish the gap between the simulator and the real world, we tested the performance of the pre-trained model on the physical husky. Unfortunately, the pre-trained TD3 model exhibits a significant decrease in performance. Our observation was that there was a slight difference of about $0.4m$ minimum distance to an obstacle returned by the physical husky when compared to that pf the virtual husky. Result of this is shown in Table \ref{tab:result}, as the success rate decreases by $16\%$ while the collision rate and timeout rate increases by $50\%$ and $6\%$ respectively. Figure \ref{fig:dt_performance} shows some of the performance of the pre-trained model in the real world environment.

\subsection{Performance with Digital Twin Retraining} 
\begin{figure*}
    \centering
    \includegraphics[width=1\linewidth]{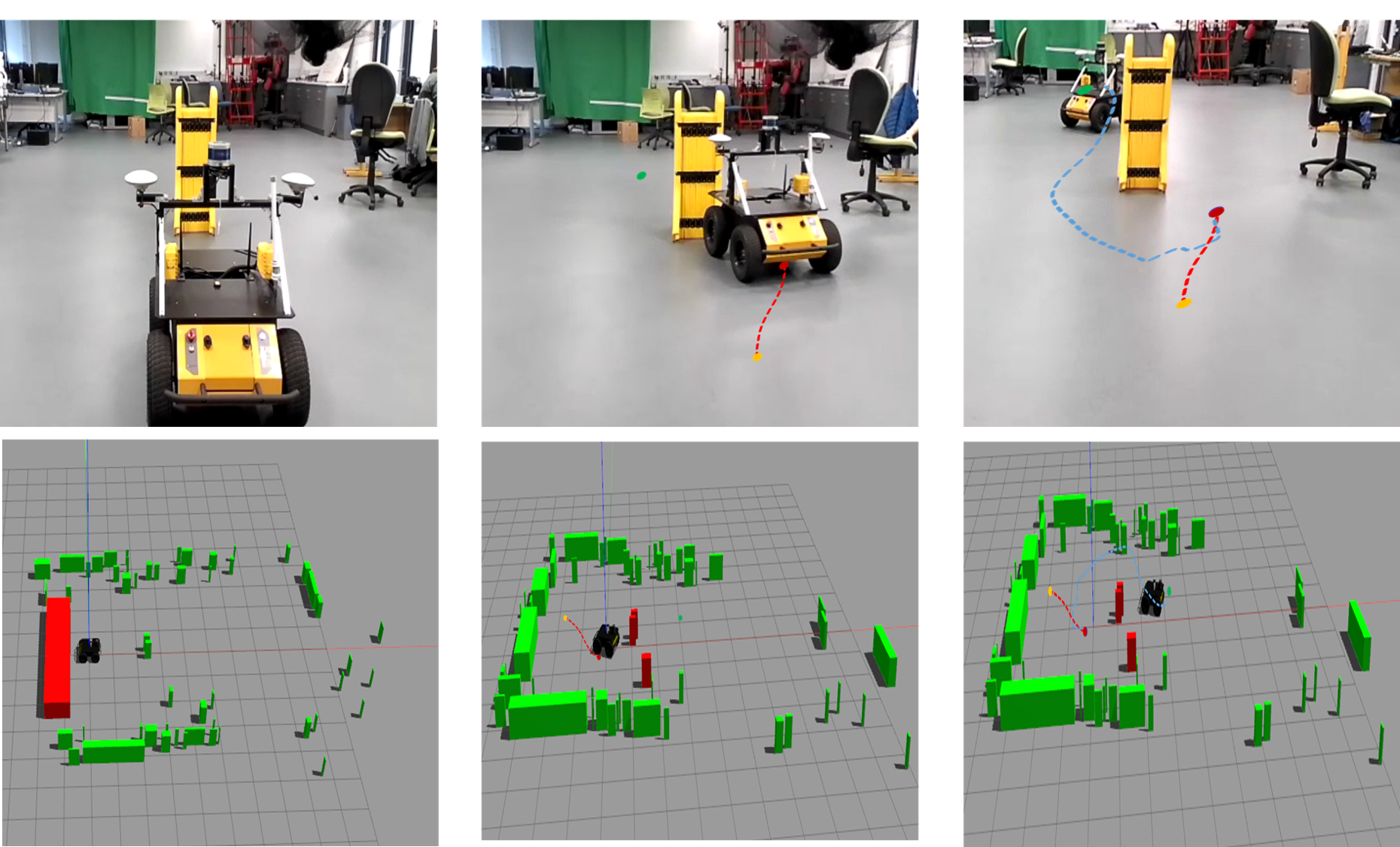}
    \caption{Illustration of the digital and virtual phases. The upper figures represent the real world state of the robot while the lower figures shows their representation in gazebo simulator. The leftmost figure shows the initial state the robot The middle figure shows the path the robot took. In the middle figure, the robot pauses because of an intention of colliding with the obstacle. The figure to the right shows the path the robot took after a retraining.}
    \label{fig:real_retraining}
\end{figure*}

As the main aim of this work is to establish retraining of the DRL model in digital twin to enhance collision avoidance, we validate the process on the real robot. We repeat the robot's navigation from a start position of 0,0 on both the x and y axes to a goal point of 5,0 on the x and y axes as in section \ref{subsec: withou_retraining}. We went further to introduce a sudden obstacle to obstruct the path the robot was on. In all of these scenarios, the robot did not collide with the obstacle rather, it pauses the movement of the physical robot, spawns the obstacle position in simulation again to enter retraining. Once a path is found, the controls are sent back to the physical robot to reach its goal, see Figure \ref{fig:real_retraining}. Our proposed method requires more navigation time to avoid collision if retraining is required. We suggest that the reader watch the accompanying video to see the visualization of these experiments.

\section{Conclusion}
\label{sec:conclusion}
In this work, we present an autonomous ground vehicle navigation using digital twin . Our approach creates the virtual twin from the LIDAR data received from the physical environment to ensure accurate representation of the physical environment. Our TD3 DRL model comprises two goals: path-planning and collision avoidance. The TD3 model is initial trained in gazebo simulator. To avoid collision during deployment on the physical robot, we propose a novel retraining of the TD3 model based on the virtual twin environment. We show that our approach is robust with an unknown environment. We also show that the virtual environment does not necessary need to be pre-designed using graphic tools. We demonstrate a real-world application in an office space.Potential enhancements for our system could involve: i) enhancing the generation of the identified obstacles in gazebo to display different shapes, ii) refining the retraining process to minimize the navigation duration, and iii) conducting experiments in a dynamic setting.

\section{Supplementary Materials:}
\textbf{GitHub: }https://github.com/kabirat/Husky-Digital-Twin
\textbf{YouTube: } https://youtu.be/EDXCZAi5aC8
\bibliographystyle{IEEEtran}
\bibliography{root}

\end{document}